\documentclass[numbers]{article}

\usepackage[final]{neurips_2024}
\usepackage{graphicx}
\usepackage{amsmath}
\usepackage{wrapfig}
\usepackage{float}
\usepackage{listings}
\usepackage{authblk}

\usepackage{multirow}
\usepackage{subcaption}
\usepackage{geometry}
\usepackage[table,xcdraw]{xcolor}
\usepackage{caption}

\usepackage[utf8]{inputenc} 
\usepackage[T1]{fontenc}    
\usepackage{hyperref}       
\usepackage{url}            
\usepackage{booktabs}       
\usepackage{amsfonts}       
\usepackage{nicefrac}       
\usepackage{microtype}      
\usepackage{xcolor}         

\title{ChemTEB: Chemical Text Embedding Benchmark, an Overview of Embedding Models Performance \& Efficiency on a Specific Domain}

\author{
\begin{minipage}[c]{\textwidth}
\centering
\textbf{
  Ali Shiraee Kasmaee\textsuperscript{1,2,*,†}, 
  Mohammad Khodadad\textsuperscript{1,2,†}, 
  Mohammad Arshi Saloot\textsuperscript{2}, 
  Nicholas Sherck\textsuperscript{3}, 
  Stephen Dokas\textsuperscript{3}, 
  Hamidreza Mahyar\textsuperscript{1}, 
  Soheila Samiee\textsuperscript{2}}\\
  \textsuperscript{1}Department of Computational Science and Engineering, McMaster University, Canada \\
  \vspace{0.2em}
  \textsuperscript{2}BASF Canada Inc., Canada\\
  \textsuperscript{3}BASF Corporation, USA\\
  \texttt{\{shiraeea, khodam3, mahyarh\}@mcmaster.ca \linebreak
  \{mohammad.arshi-saloot, nicholas.sherck, stephen.dokas, soheila.samiee\}@basf.com} \\
  \textsuperscript{*}Corresponding Author: \texttt{shiraeea@mcmaster.ca}\\
    \textsuperscript{†}Equal Contribution
\end{minipage}
}

\begin{document}
\maketitle

\begin{abstract}
Recent advancements in language models have started a new era of superior information retrieval and content generation, with embedding models playing an important role in optimizing data representation efficiency and performance. While benchmarks like the Massive Text Embedding Benchmark (MTEB) have standardized the evaluation of general domain embedding models, a gap remains in specialized fields such as chemistry, which require tailored approaches due to domain-specific challenges. 
This paper introduces a novel benchmark, the Chemical Text Embedding Benchmark (ChemTEB), designed specifically for the chemical sciences. ChemTEB addresses the unique linguistic and semantic complexities of chemical literature and data, offering a comprehensive suite of tasks on chemical domain data. 
Through the evaluation of 34 open-source and proprietary models using this benchmark, we illuminate the strengths and weaknesses of current methodologies in processing and understanding chemical information. Our work aims to equip the research community with a standardized, domain-specific evaluation framework, promoting the development of more precise and efficient NLP models for chemistry-related applications. Furthermore, it provides insights into the performance of generic models in a domain-specific context. 
ChemTEB comes with open-source code \footnote{\href{https://github.com/basf/chemteb}{https://github.com/basf/chemteb}} and data \footnote{\href{https://huggingface.co/BASF-AI}{https://huggingface.co/BASF-AI}}, contributing further to its accessibility and utility.
\end{abstract}

\section{Introduction}
Deep learning and natural language processing have improved significantly, highlighting the importance of learning text representation to understand semantic similarity. This is a key part of text mining, search, retrieval, and other similar tasks. In the last decade, many promising models have been developed to meet this need. Earlier models, such as GloVe \cite{pennington2014glove} and Word2vec \cite{mikolov2013efficient} focused on word embedding but lacked context awareness. More recent models have adopted transformer architecture to incorporate context into token embeddings \cite{vaswani2017attention}. Each model proposes unique architectural features, parameter counts, context lengths, and pretraining methods. 

BERT was among the first models to employ transformer architecture and self-supervised training for embedding models \cite{vaswani2017attention, devlin2018bert}. Subsequently, variants of BERT were introduced to enhance performance, such as ROBERTA \cite{liu2019roberta}, or to facilitate domain adaptation for specific areas of interest, such as SciBERT for the sciences \cite{beltagy2019scibert}. 
Initially, a common approach to achieve a single embedding vector for a text corpus -- i.e. pooling -- was to average the output layer or use the [CLS] token; however, it has been found to be less effective in grasping semantically meaningful embeddings. Sentence-BERT \cite{reimers2019sentence} is a promising fine-tuned variant that leverages a Siamese bi-encoder and triplet loss \cite{schroff2015facenet} to achieve competitive performance in semantic representation learning, making it highly suitable for embedding tasks. Models such as E5 \cite{wang2022text} and Nomic embed \cite{nussbaum2024nomic}  have integrated contrastive learning into the pretraining process in an effort to help the models better distinguish between similar and dissimilar samples, thereby enhancing implementation efficiency. The BGE model family \cite{xiao2024c} stands out by integrating an MAE-inspired pre-training approach \cite{xiao2022retromae} with contrastive learning, utilizing large batch sizes to enhance embedding quality.  M3-embedding \cite{multim3} has emphasized multi-functionality, multi-granularity, and multi-linguality in its embedding model to further improve the performance. Furthermore, OpenAI, Cohere, and Amazon have introduced their proprietary models for embedding, continually expanding the list of available models. 

The progress in natural language processing (NLP) has influenced a wide range of scientific domains, including biology, medicine, and physics. These advancements have enabled researchers to extract, analyze, and interpret vast amounts of textual data with unprecedented accuracy and efficiency. Embedding models are crucial for solving complex tasks across these domains. These models transform high-dimensional data into dense vector spaces, capturing semantic relationships essential for applications such as chemical literature mining and even molecular property prediction. 
The rise of Retrieval-Augmented Generation (RAG) architectures \cite{lewis2020retrieval}, which combine language models with external knowledge retrieval systems, offers new opportunities for these tasks. 
RAG enhances embedding-based applications by allowing dynamic access to domain-specific knowledge, making them highly effective for tasks that require both deep learning and external information sources. With the increased application of such methods, the requirement of having efficient embedding models is raised in the industry. 
While general NLP benchmarks like the Massive Text Embedding Benchmark (MTEB) have been instrumental in standardizing model evaluations across a variety of tasks \cite{muennighoff2022mteb}, they fall short when applied to chemistry-specific tasks. For example, the linguistic and semantic nuances inherent in chemical literature are often overlooked by models trained and evaluated on general datasets. This gap underscores the need for a specialized benchmark tailored to the domain of chemistry, where precision and context are of paramount importance.
Improved NLP models have the potential to revolutionize various aspects of chemistry, such as automated literature reviews, chemical synthesis planning, patent analysis and even contribute to further improvement of Autonomous Agents in Chemistry \cite{bran2023chemcrow, ramos2024review}.

To bridge this gap, we introduce a novel benchmark explicitly tailored for the chemical sciences, named Chemical Text Embedding Benchmark (ChemTEB). This benchmark offers a comprehensive suite of tasks, ranging from chemical text classification to bitext mining of natural language and SMILES representation of chemical compounds. 
 Our benchmark aims to provide a robust, domain-specific evaluation framework, promoting the development of more precise and efficient NLP models for applications related to chemistry. Accompanied by an open-source package and data, ChemTEB enables straightforward evaluation of any model and facilitates the effortless incorporation of new tasks and datasets.

\section{Related Work}
To the best of our knowledge, there is no previous benchmark on chemical embedding models. The closest previous benchmarks were either generic NLP benchmarks on information retrieval or semantic similarity; or chemical benchmarks on generative large language models, machine learning methods on molecular data or Graph Neural Networks (GNNs). In the following, more details on these benchmarks are provided.  

\subsection{NLP Benchmarks}
The evaluation of sentence embeddings and general language-understanding has seen significant advancements, driven by various benchmarks and toolkits designed to assess models across a diverse set of tasks. One prominent toolkit is SentEval \cite{conneau2018senteval}, widely used for evaluating sentence embeddings on tasks such as semantic textual similarity, paraphrase detection, and sentiment analysis. SentEval is complemented by the GLUE Benchmark \cite{wang2018glue}, which assesses models on a broader range of natural language processing (NLP) tasks, including sentence embeddings, to measure language understanding. Building on this success, SuperGLUE \cite{wang2019superglue} takes evaluation a step further, introducing more complex tasks that challenge models to push beyond existing performance levels. For more targeted evaluation, the STS Benchmark \cite{cer2017semeval} focuses on measuring semantic similarity between sentence pairs, which is critical for determining the quality of sentence embeddings. In the context of information retrieval, BEIR \cite{thakur2021beirheterogenousbenchmarkzeroshot} provides a heterogeneous benchmark, incorporating datasets that evaluate models on tasks like retrieval, ranking, and embedding quality.

Cross-lingual performance is another essential dimension in embedding evaluation, tackled by benchmarks like XTREME \cite{hu2020xtreme}, which tests models on classification, retrieval, and sentence similarity tasks across multiple languages. XGLUE \cite{liang2020xglue}, a multilingual extension of the GLUE framework, offers further evaluation on a wide range of NLP tasks across different languages. Meanwhile, the TREC benchmarks \cite{soboroff2021overview} have been a long-standing standard for testing information retrieval systems, particularly for embedding models in retrieval scenarios. MTEB \cite{muennighoff2022mteb} broadens the scope by evaluating embeddings across a variety of language understanding tasks and domains, focusing on scalability and adaptability. Additionally, specialized NLP tasks like natural language inference and paraphrase recognition are tackled by datasets such as MultiNLI \cite{kim2019semantic} and PAWS \cite{zhang2019paws}, which offer challenging benchmarks for testing the depth of understanding in sentence embeddings. For Chinese NLP tasks, CLUE \cite{xu2020clue} serves as a counterpart to GLUE, addressing the unique challenges of the Chinese language. Despite the breadth of these benchmarks, one notable gap remains; the absence of a comparable evaluation framework for chemistry-related language models.

\subsection{Chemistry Benchmarks}
While there have been attempts to apply NLP in chemistry, such as ChemDataExtractor \cite{swain2016chemdataextractor} and natural language processing techniques to extract information on the properties and functionalities of energetic materials from large text corpora \cite{elton2019using}, these efforts are often limited by the lack of a standardized evaluation framework. Existing resources like the ChEMBL database \cite{zdrazil2024chembl} offer valuable data but do not provide the comprehensive, task-oriented benchmarking necessary to drive significant advances in chemical NLP. 
The ChemNLP \cite{choudhary2023chemnlp} library advances the application of NLP in chemistry by providing curated datasets from arXiv and PubChem, along with tools for visualization, analysis, and task execution tailored to materials chemistry. However, there is no proper task or code for the evaluation of developed models.  

After the rise of large generative models, some benchmark datasets and evaluation frameworks have been developed to assess the capabilities of large language models (LLMs) and machine learning techniques in the fields of materials science and chemistry. MaterialBENCH \cite{yoshitake2024materialbenchevaluatingcollegelevelmaterials}, ChemBench  \cite{mirza2024largelanguagemodelssuperhuman}, and ChemLLMBench \cite{guo2023gpt} are some of these studies. However, none of them provided tasks and data on \textit{embedding} models evaluation.

\section{ChemTEB}
In this work, we leverage a diverse set of datasets collected to evaluate embedding models across diverse tasks, including Classification, Pair Classification, Clustering, Retrieval, and Bitext Mining. The data sources used are PubChem \cite{kim2023pubchem}, English Wikipedia, BeIR \cite{thakur2021beir}, CoconutDB \cite{sorokina2021coconut}, and Safety Data Sheets \cite{pereira2020msds} each offering unique and complementary information critical for evaluating the performance of NLP models in chemistry. All tasks and datasets are either designed or validated by domain experts, i.e. chemists.

\subsection{Tasks}
\label{sec:tasks}
We present a variety of benchmarks designed to evaluate different aspects of natural language and chemical data processing. Each benchmark focuses on a specific task, and in this section, we provide an overview of the task, the data sources used for its collection, and the evaluation process. These benchmarks provide a comprehensive set of evaluation tasks using diverse datasets tailored to different modeling approaches. Table~\ref{tab:datasets-summary} provides a summary of datasets and statistics associated with them.

\begin{table}[ht!]
	\centering
	\caption{Datasets summary. This table provides an overview of the datasets used across different tasks, including the dataset names from Hugging Face, the original data sources, and the distribution of sample sizes. The distribution is represented through key statistical measures: 5th percentile, median, and 95th percentile of the number of tokens }
	\resizebox{\textwidth}{!}{
		\begin{tabular}{llllllll}
			\hline
			& \multicolumn{1}{c}{}                   & \multicolumn{1}{c}{}                                                     & \multicolumn{1}{c}{}                              & \multicolumn{1}{c}{}                                                                                      & \multicolumn{3}{c}{Sequence Lengths (tokens \footnotemark)} \\
   \cmidrule{6-8}
			\multirow{-2}{*}{Task}               & \multicolumn{1}{c}{\multirow{-2}{*}{}} & \multicolumn{1}{c}{\multirow{-2}{*}{HuggingFace Name}}                   & \multicolumn{1}{l}{\multirow{-2}{*}{Data Source}} & \multicolumn{1}{c}{\multirow{-2}{*}{\#Samples}} & 5th Percentile & Median & 95th Percentile  \\
			\hline \\
			                                     & 1                                      & WikipediaEasy10Classification                                            & Wikipedia                                         & 2105                                            & 42  & 178    & 612.4 \\
			                                     & 2                                      & WikipediaEasy5Classification                                             & Wikipedia                                         & 1164                                            & 43  & 171.5  & 547.85 \\
			                                     & 3                                      & WikipediaMedium5Classification                                           & Wikipedia                                         & 617                                             & 39  & 137    & 563.6 \\
			                                     & 4                                      & WikipediaMedium2CrystallographyVsChromatographyTitrationpHClassification & Wikipedia                                         & 1451                                            & 41.5  & 175    & 658.5 \\
			                                     & 5                                      & WikipediaMedium2BioluminescenceVsNeurochemistryClassification            & Wikipedia                                         & 486                                             & 42  & 158    & 574.25 \\
			                                     & 6                                      & WikipediaEZ2Classification                                               & Wikipedia                                         & 58921                                           & 41  & 164    & 590 \\
			                                     & 7                                      & WikipediaHard2BioluminescenceVsLuminescenceClassification                & Wikipedia                                         & 410                                             & 41  & 148.5  & 579.3 \\
			                                     & 8                                      & WikipediaEasy2GeneExpressionVsMetallurgyClassification                   & Wikipedia                                         & 5741                                            & 42  & 175    & 630 \\
			                                     & 9                                      & WikipediaEasy2GreenhouseVsEnantiopureClassification                      & Wikipedia                                         & 1136                                            & 34  & 139.5  & 513 \\
			                                     & 10                                     & WikipediaEZ10Classification                                              & Wikipedia                                         & 43146                                           & 41  & 165    & 582 \\
			                                     & 11                                     & WikipediaHard2SaltsVsSemiconductorMaterialsClassification                & Wikipedia                                         & 491                                             & 38.5  & 141    & 447.5 \\
			                                     & 12                                     & WikipediaEasy2SolidStateVsColloidalClassification                        & Wikipedia                                         & 2216                                            & 42  & 151    & 532 \\
			                                     & 13                                     & WikipediaMedium2ComputationalVsSpectroscopistsClassification             & Wikipedia                                         & 1101                                            & 38  & 155    & 639 \\
			                                     & 14                                     & WikipediaHard2IsotopesVsFissionProductsNuclearFissionClassification      & Wikipedia                                         & 417                                             & 43.8  & 209    & 706.4 \\
			\multirow{-15}{*}{Classification}    & 15                                     & WikipediaEasy2SpecialClassification                                      & Wikipedia                                         & 1312                                            & 35.55  & 133    & 465 \\
			                                     & 16                                     & SDSGlovesClassification                                                  & Safety Data Sheets                                & 8000                                            & 498  & 1071   & 1871 \\
			                                     & 17                                     & SDSEyeProtectionClassification                                           & Safety Data Sheets                                & 8000                                            & 492 & 1060   & 1876 \\
                                        \cmidrule{2-8}
			                                     & 18                                     & CoconutSMILES2FormulaBM                                          & CoconutDB                                         & 8000                                           & 6   & 11     & 150\\
			                                     & 19                                     & PubChemSMILESISoTitleBM                                                  & PubChem                                           & 14140                                           & 4   & 22     & 93 \\
			                                     & 20                                     & PubChemSMILESISoDescBM                                                   & PubChem                                           & 14140                                           & 12   & 45     & 134 \\
			                                     & 21                                     & PubChemSMILESCanonTitleBM                                                & PubChem                                           & 30914                                           & 3   & 12     & 43 \\
			\multirow{-6}{*}{BitextMining}   & 22                                     & PubChemSMILESCanonDescBM                                                 & PubChem                                           & 30914                                           & 8   & 24     & 109 \\
			                            \cmidrule{2-8}        
                                        & 23                                     & ChemHotpotQARetrieval                                                    & HotpotQA                                          & 10275                                           & 19   & 71          & 183 \\
			\multirow{-2}{*}{Retrieval}          & 24                                     & ChemNQRetrieval                                                          & Natural Questions                                  & 22960                                           & 13   & 81          & 231 \\
			                         \cmidrule{2-8}          
                                        & 25                                     & WikipediaMedium5Clustering                                               & Wikipedia                                         & 617                                             & 39  & 137    & 563.6\\
			\multirow{-2}{*}{Clustering}         & 26                                     & WikipediaEasy10Clustering                                                & Wikipedia                                         & 2105                                            & 42  & 178    & 612.4 \\
			                           \cmidrule{2-8}          
                                        & 27                                     & WikipediaAIParagraphsParaphrasePC                                          & Wikipedia                                           & 5408                                             & 28  & 104    & 354 \\
			                                     & 28                                     & CoconutSMILES2FormulaPC                                                     & CoconutDB                                         & 8000                                         & 6   & 11     & 108 \\
			                                     & 29                                     & PubChemAISentenceParaphrasePC                    & PubChem                                           & 4096                                            & 9   & 20     & 59 \\
			                                     & 30                                     & PubChemSMILESCanonTitlePC                        & PubChem                                           & 4096                                            & 4   & 16     & 30 \\
			                                     & 31                                     & PubChemSynonymPC                                 & PubChem                                           & 4096                                            & 3   & 8      & 38 \\
			                                     & 32                                     & PubChemSMILESCanonDescPC                         & PubChem                                           & 4096                                            & 12   & 23     & 105 \\
			                                     & 33                                     & PubChemSMILESIsoDescPC                           & PubChem                                           & 4096                                            & 12   & 48     & 125 \\
			                                     & 34                                     & PubChemSMILESIsoTitlePC                          & PubChem                                           & 4096                                            & 4   & 35     & 70 \\
			\multirow{-9}{*}{PairClassification} & 35                                     & PubChemWikiParagraphsPC                          & PubChem                                           & 4096                                            & 8   & 66     & 235 \\  \\
			\hline
		\end{tabular}
}
\label{tab:datasets-summary}
\end{table}
\footnotetext{All tokenizations are done with OpenAI's \textit{cl100k\_base} tokenizer via tiktoken python package.}

\textbf{Classification}: Each task consists of a dataset with a textual field and associated labels. A logistic regression classifier is used on top of an embedding model, which is first trained on the training split of the dataset. The performance is then evaluated on the test dataset using the F1 score. We have used two main sources to construct the datasets in this category: first, chemistry-related English Wikipedia articles classified into various chemistry subfields, and second, Safety Data Sheets (SDS) \cite{pereira2020msds}, which are detailed documents providing essential information on the properties and hazards of chemicals, ensuring user safety and compliance with regulatory standards.

\textbf{Clustering} tasks involve grouping related text pieces into meaningful clusters based on their embeddings. Similar to the classification tasks, clustering datasets are also constructed from chemistry-related English Wikipedia articles, where section texts are clustered into various chemistry subfields. A mini-batch k-means model with a batch size of 32 is trained on the text pieces. We have used V-measure \cite{rosenberg2007v} to assess and report the performance.

\textbf{Pair classification} tasks involve determining whether two textual samples are related and assigning a binary label to them. In the chemistry domain, this relationship could be whether two texts refer to the same chemical entity, or matching compound titles or their descriptions with their SMILES strings (Simplified Molecular Input Line Entry System). In this task, a pair of text is embedded using an embedding model, their distance is calculated (using metrics such as cosine similarity, Euclidean distance, Manhattan distance and dot product), and the best binary threshold for each metric is determined. The F1 score is then calculated across all metrics, with the maximum F1 score reported as the main score. The datasets for these tasks are constructed from databases containing chemical product information, such as PubChem \cite{kim2023pubchem} and COCONUT \cite{sorokina2021coconut}.

\textbf{Bitext Mining} focuses on matching pairs of text that are translations or semantic equivalents of each other, by performing a semantic similarity search between a list of query embeddings and a list of corpus embeddings. We have used data sources such as PubChem \cite{kim2023pubchem} and COCONUT \cite{sorokina2021coconut} and matched the SMILES of chemical entities with their titles, descriptions and chemical formulas. We have employed f1 score to evaluate the performance of the models. 
 
In \textbf{Retrieval} tasks, the model's ability to retrieve relevant records based on a query is evaluated. Each dataset in this category consists of a list of queries and documents, along with the mapping between relevant queries and documents. A model is used to embed all the texts, and the relevant documents are retrieved based on the cosine similarity between embeddings. We have used a chemistry-related subset of the Natural Questions \cite{kwiatkowski2019natural} and HotpotQA \cite{yang2018hotpotqa} datasets and employed nDCG@10 as the main evaluation metric.

\subsection{Embedding Models}
In this study, a total of 34 embedding models have been evaluated in the ChemTEB benchmark. This includes 27 open-source models and 7 proprietary models. More detailed information about the evaluated models can be found in section \ref{section:supp_embed_models} of supplementary materials, with Table~\ref{tab:models-summary} comparing these models based on their characteristics.

\subsection{Ranking Process for Model Performance}
\label{section:rank}

The models are ranked based on their performance across datasets in each category of tasks.
In the first step, the arithmetic mean of the performance metrics is computed for each task to summarize the performance over tasks. 

In the next step, an overall score was calculated using the Reciprocal Rank Fusion (RRF) method \cite{cormack2009reciprocal}:
\[
\text{RRF}_{\text{score}}(m) = \sum_{d \in \text{Datasets}} \frac{1}{k + r_d(m)}
\]

where $r_d(m)$ is the rank of model $m$ for dataset $d$, and $k = 10$ ensures all models retain weight. Summing RRF scores across datasets gives an aggregate score for each model. Higher RRF scores reflect a better overall ranking of the model.

\section{Results}
\subsection{Models Performance}
Table~\ref{tab:summary_performance} summarizes the average performance of each model in each category of tasks and the overall performance, which is presented with \(RRF_{score}\) (see section~\ref{section:rank} for more details).
From \textbf{model} perspective, there is no single model which outperforms others in all tasks, but in general proprietary models provide a better performance compared to open-source models. \textit{OpenAI-text embedding 3-large} provided the best results in 3 out of 5 task categories, ranked first among evaluated models. Among open-source models, \textit{Nomic Embedding v1.5} showed the best overall performance, ranking second after OpenAI large embedding model (see the detailed ranking of all models on each sub-task in Supplementary Table~\ref{tab:summary_rank}).

\begin{table}

\centering
\caption{Summary of models performance. This table provides a comprehensive comparison of models across several key tasks: text classification (macro F1-score), bitext mining (F1-score), text retrieval (nDCG@10), text clustering (F1-score), pair classification (maximum F1-score), and an overall score (Reciprocal Rank Fusion). The models are grouped into two categories—open-source and proprietary—for easier distinction. In each category, the best-performing model is underscored, while the overall best-performing model across all categories is highlighted in bold. }
\resizebox{\textwidth}{!}{
\begin{tabular}{lllllll}
\toprule
 & Classification & Bitext Mining & Retrieval & Clustering & Pair Classification & Final Score\\
 & (Macro F1) & (F1) & (nDCG@10) & (V-measure) & (Max F1)  & (RRF) \\
\midrule
BERT & 0.72±0.04 & 0.0±0.0 & 0.28±0.02 & 0.2±0.03 & 0.41±0.05 & 0.122 \\
SciBERT & 0.71±0.04 & 0.0002±0.0 & 0.2±0.03 & 0.18±0.02 & 0.43±0.05 & 0.122 \\
MatSciBERT & 0.7±0.04 & 0.0003±0.0001 & 0.11±0.02 & 0.21±0.03 & 0.41±0.05 & 0.122 \\
Chemical BERT & 0.68±0.04 & 0.0003±0.0 & 0.17±0.01 & 0.13±0.02 & 0.42±0.05 & 0.120 \\
Nomic BERT & 0.67±0.04 & 0.0001±0.0 & 0.05±0.0 & 0.22±0.03 & 0.38±0.04 & 0.118 \\
Nomic Embedding v1 & 0.77±0.04 & 0.0023±0.0002 & 0.72±0.02 & 0.46±0.03 & \textbf{\underline{0.55±0.06}} & 0.285 \\
Nomic Embedding v1.5 & 0.78±0.04 & 0.0026±0.0002 & 0.75±0.02 & 0.5±0.04 & \textbf{\underline{0.55±0.06}} & \underline{0.339} \\
SBERT - all Mini LM L6.v2 & \underline{0.78±0.03} & 0.0015±0.0002 & 0.61±0.01 & 0.36±0.02 & 0.54±0.06 & 0.232 \\
SBERT - all Mini LM L12.v2 & 0.77±0.04 & 0.0013±0.0001 & 0.58±0.0 & 0.34±0.01 & 0.54±0.06 & 0.201 \\
SBERT - all MPNET-base.v2 & 0.78±0.04 & 0.001±0.0001 & 0.56±0.0 & 0.5±0.03 & 0.54±0.06 & 0.239 \\
SBERT - multi-qa-mpnet-base.v1 & 0.74±0.04 & 0.0009±0.0001 & 0.56±0.01 & 0.42±0.04 & 0.54±0.06 & 0.185 \\
E5 - small & 0.75±0.03 & 0.0015±0.0001 & 0.69±0.02 & 0.12±0.02 & 0.48±0.05 & 0.166 \\
E5 - base & 0.76±0.04 & 0.0019±0.0001 & 0.68±0.01 & 0.34±0.05 & 0.49±0.05 & 0.192 \\
E5 - large & 0.77±0.04 & \underline{0.0029±0.0002} & 0.7±0.01 & \underline{0.51±0.04} & 0.5±0.05 & 0.290 \\
E5 - small v2 & 0.76±0.03 & 0.0012±0.0001 & 0.69±0.01 & 0.19±0.03 & 0.46±0.05 & 0.165 \\
E5 - base v2 & 0.76±0.04 & 0.0016±0.0001 & 0.68±0.01 & 0.38±0.05 & 0.47±0.05 & 0.178 \\
E5 - large v2 & 0.76±0.04 & 0.0022±0.0002 & 0.73±0.01 & 0.33±0.05 & 0.48±0.05 & 0.214 \\
E5 - Multilingual small & 0.74±0.04 & 0.0018±0.0001 & \textbf{\underline{0.76±0.01}} & 0.17±0.01 & 0.47±0.05 & 0.207 \\
E5 - Multilingual base & 0.75±0.04 & 0.0022±0.0001 & 0.68±0.0 & 0.48±0.03 & 0.47±0.05 & 0.196 \\
E5 - Multilingual large & 0.74±0.04 & 0.0026±0.0002 & 0.67±0.0 & 0.3±0.05 & 0.48±0.05 & 0.187 \\
BGE - small en & 0.78±0.04 & 0.0012±0.0001 & 0.52±0.04 & 0.27±0.03 & 0.48±0.05 & 0.160 \\
BGE - base en & 0.77±0.04 & 0.0019±0.0001 & 0.59±0.03 & 0.44±0.05 & 0.48±0.05 & 0.186 \\
BGE - large en & 0.78±0.04 & 0.0016±0.0001 & 0.44±0.06 & 0.45±0.05 & 0.49±0.05 & 0.191 \\
BGE - small en v1.5 & \underline{0.78±0.03} & 0.0013±0.0001 & 0.63±0.03 & 0.25±0.04 & 0.48±0.05 & 0.180 \\
BGE - base en v1.5 & 0.77±0.04 & 0.0018±0.0001 & 0.69±0.02 & 0.47±0.05 & 0.49±0.05 & 0.219 \\
BGE - large en v1.5 & 0.78±0.04 & 0.0019±0.0001 & 0.67±0.02 & 0.39±0.06 & 0.5±0.05 & 0.224 \\
BGE - Multilingual - M3 & 0.76±0.03 & 0.0012±0.0002 & 0.68±0.02 & 0.45±0.05 & 0.47±0.06 & 0.176 \\
\cmidrule{2-7}
OpenAI - Text embedding 3 - small & 0.78±0.04 & 0.0027±0.0003 & 0.65±0.01 & 0.49±0.05 & 0.5±0.05 & 0.273 \\
OpenAI - Text embedding 3 - large & 0.8±0.04 & \textbf{\underline{0.0062±0.0006}} & \underline{0.71±0.01} & \textbf{\underline{0.6±0.03}} & \underline{0.53±0.05} & \textbf{\underline{0.384}} \\
OpenAI - Text embedding - Ada - 02 & 0.78±0.04 & 0.0035±0.0002 & 0.66±0.02 & 0.52±0.04 & 0.49±0.05 & 0.279 \\
Amazon - Titan Text Embedding v2 & 0.77±0.03 & 0.0024±0.0002 & 0.62±0.0 & 0.49±0.04 & 0.49±0.05 & 0.224 \\
Amazon - Titan Embedding G1 Text & \textbf{\underline{0.81±0.03}} & 0.0032±0.0003 & 0.6±0.02 & 0.45±0.06 & 0.49±0.05 & 0.285 \\
Cohere - Embed English V3 & \textbf{\underline{0.81±0.03}} & 0.0012±0.0 & 0.49±0.04 & 0.55±0.02 & \underline{0.53±0.06} & 0.278 \\
Cohere - Embed Multilingual V3 & 0.8±0.03 & 0.0024±0.0001 & 0.49±0.04 & 0.53±0.03 & \underline{0.53±0.06} & 0.281 \\
\bottomrule
\end{tabular}
}
\label{tab:summary_performance}
\end{table}

\begin{figure}[ht!]
    \centering
    \includegraphics[width=\textwidth]{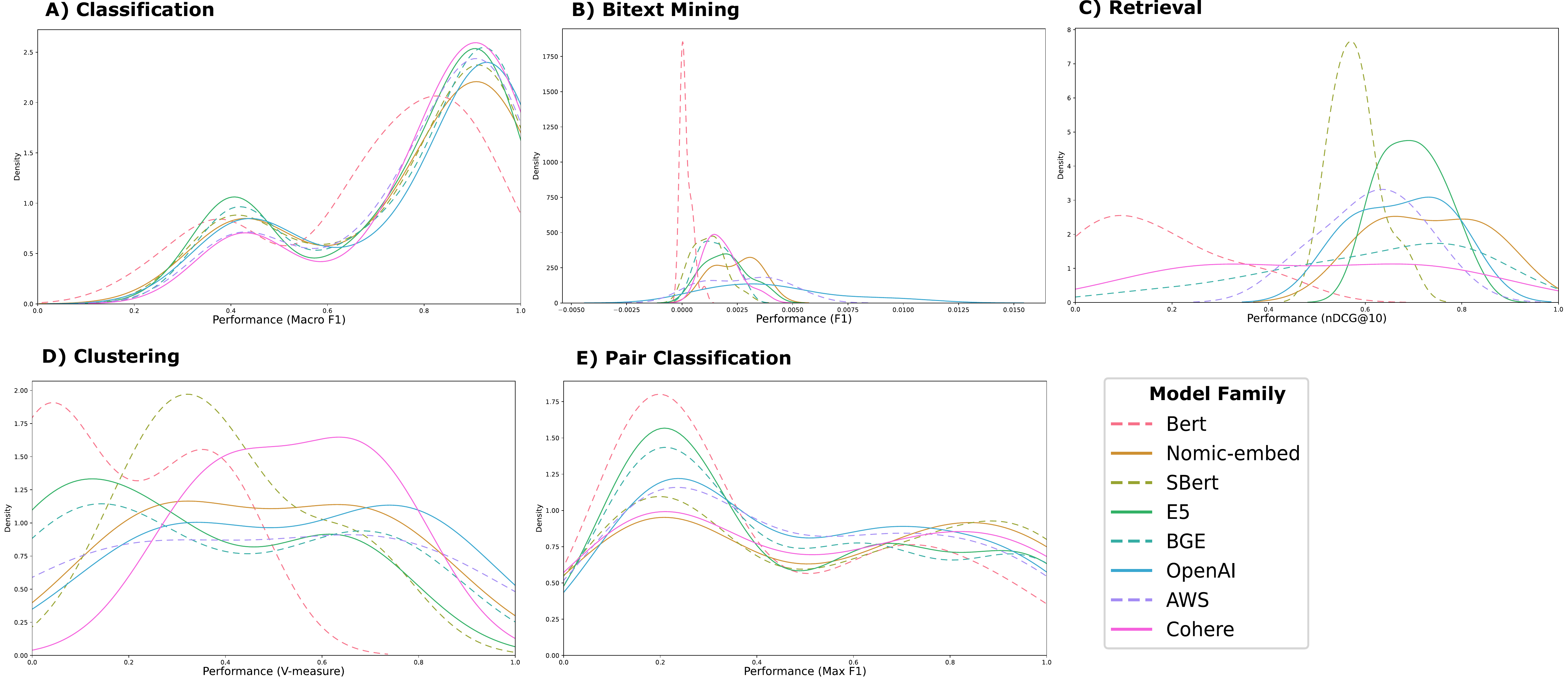}
    \caption{Distribution plots for five categories of tasks. The KDE plots show the probability density functions, where the x-axis represents the range of predicted values (performance distribution over tasks of each category and models of each family) and the y-axis represents the estimated density. Each colored line corresponds to a unique model family, enabling a clear visual comparison of their value distributions.}
    \label{fig:performance_histogram}
\end{figure}

From \textbf{Task} category perspective, models generally performed better in classification tasks, and the worst performance has been observed in bitext mining. As explained earlier in section~\ref{sec:tasks} bitext mining was designed on translation between SMILES representation of chemical compounds and either their title or description. 
The poor performance of the models in this task can be attributed to the fact that none of the general-purpose models are trained on a modality such as SMILES code. As a result, they fail to grasp the semantic relationships between different SMILES strings, leading to suboptimal performance in this specific task.
Retrieval, clustering, and pair classification, were in 2nd to 4th rank respectively, in terms of difficulty level and overall performance of models.

The models evaluated in this benchmark share many features, such as architecture and training methods. For better evaluation of the impact of models' characteristics in their performance in each category of tasks, we grouped these models based on their similarities into eight families: (i) BERT Family, (ii) Nomic embedding family, (iii) SBERT family, (ix) E5 family, (x) BGE family, (xi) OpenAI family, (xii) Amazon family, and (xiii) Cohere family. 
Figure~\ref{fig:performance_histogram} illustrates the distribution of each model family performance over all datasets in each category of tasks using the Kernel Density Estimation (KDE - see supplementary section~\ref{sec:sup_KDE} for more details).

\subsection{Models Efficiency}
\label{section:model_efficiency}
\begin{figure}[ht!]
    \centering
    \includegraphics[width=.75\textwidth]{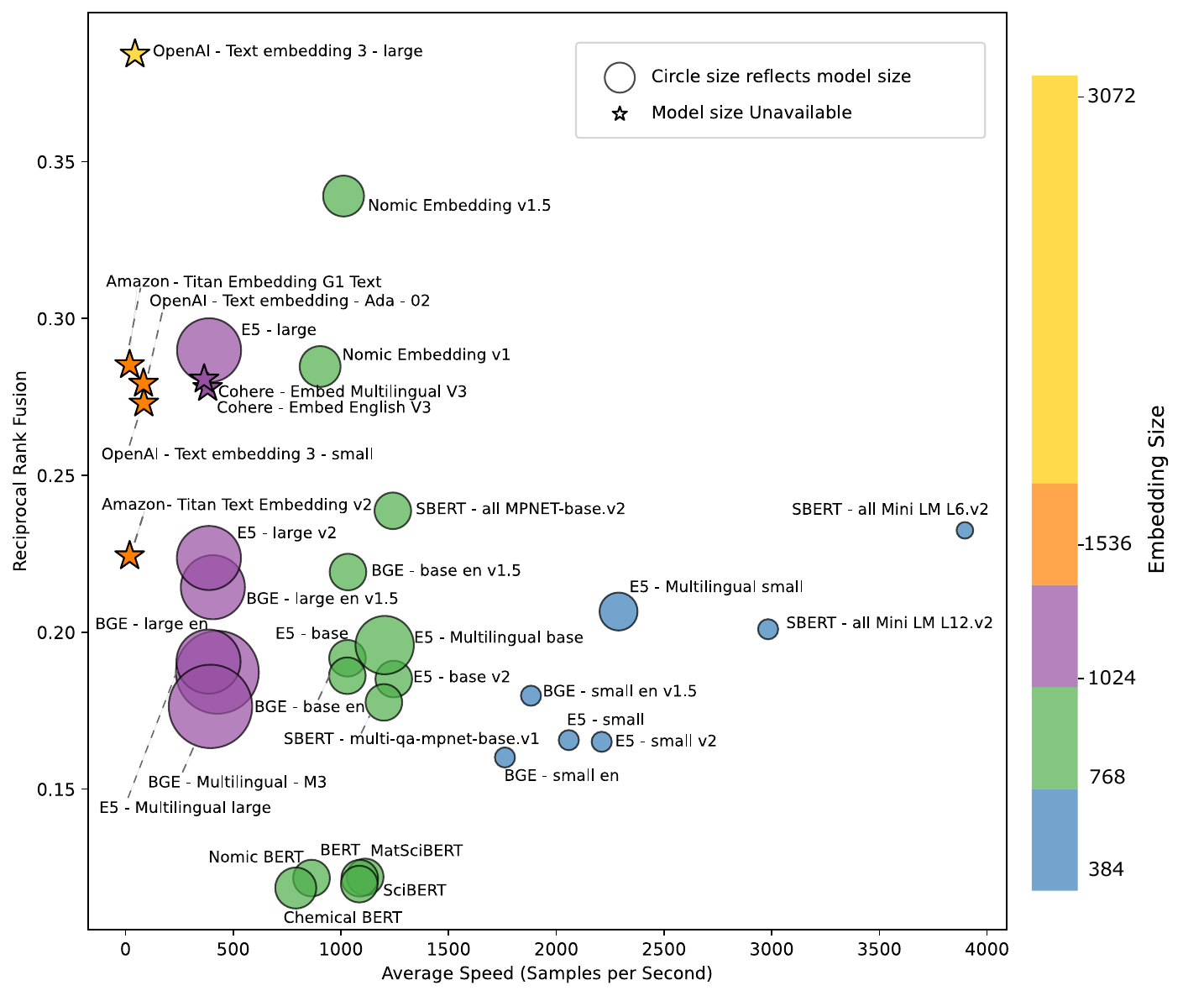}
    \caption{Summary of evaluated models in terms of efficiency. All evaluated models are depicted in the form of (i) circles (with circle size being proportional to the number of parameters) for open-source models, and (ii) stars for proprietary models. The color of the depicted models reflects their embedding dimension. The x-axis denotes the averaged inference speed (embedded samples/sec) calculated over seven pair classification tasks (tasks 29 - 35 in table~\ref{tab:datasets-summary}) conducted on a V100 GPU machine.}
    \label{fig:speed_size_RRF}
\end{figure}

The evaluated models in this benchmark vary in terms of architecture, data size, model size, speed, and performance, among other factors. Depending on specific requirements, one model may be more suitable than another, but making an informed decision requires comparing models across multiple features simultaneously. To facilitate this, in Figure~\ref{fig:speed_size_RRF}, we visualized the speed (in the pair classification task), model size, embedding size, and RRF score for each model. For proprietary models, size information was not available.

The visualization highlights the diverse performance of the models, each with its own advantages and disadvantages. A noticeable trend is that slower models tend to be larger, with bigger embedding sizes, and generally offer higher performance. On one end of the spectrum, \textit{OpenAI - Text Embedding 3 - Large} achieved the highest RRF score but exhibited very low speed. In contrast, \textit{SBERT - All Mini LM L6.v2} was both the smallest and fastest model, though with lower performance. Interestingly, \textit{BERT-based} models had the lowest RRF scores and relatively slower speeds, clearly distinguishing them from other models. Notably, the open-source \textit{Nomic Embedding v1.5} demonstrated a good balance between high speed and strong performance.

\subsection{Domain Adaptation}
To the best of our knowledge, the only existing embedding models specifically adapted to the chemical domain are MatSciBERT \cite{gupta2022matscibert}, and ChemicalBERT (from Recobo\footnote{\href{https://huggingface.co/recobo/chemical-bert-uncased}{recobo/chemical-bert-uncased}}). SciBERT \cite{beltagy2019scibert}, pre-trained on scientific data, is also more relevant to the chemistry domain compared to generic models. Within the BERT family, these domain-adapted models outperformed BERT-base in the bitext mining task, the most challenging in-domain task requiring partial knowledge of SMILES codes. However, we did not observe a consistent significant improvement in other tasks, except for SciBERT's superior performance in pair classification.  

Contrarily, outside the BERT family, these domain-adapted models performed considerably worse in most task categories, as evidenced by their joint lowest RRF-score for ranking (refer to the Table~\ref{tab:summary_rank} in supplementary materials for their detailed ranking per task category). This could be attributed to their belonging to the BERT Family with only Masked Language Modeling (MLM) pretraining. Based on the evaluations and observations, it appears that the contrastive objective and architectural improvements introduced post-BERT for enhanced semantic representation contribute more to the performance in specific domains than domain adaptation of weaker architectures. This finding encourages the research community to move away from relying on older domain-adapted models and instead continue developing domain-specific models with newer more efficient architectures.

\begin{figure}[H]
    \centering
    \includegraphics[width=0.48
    \textwidth]{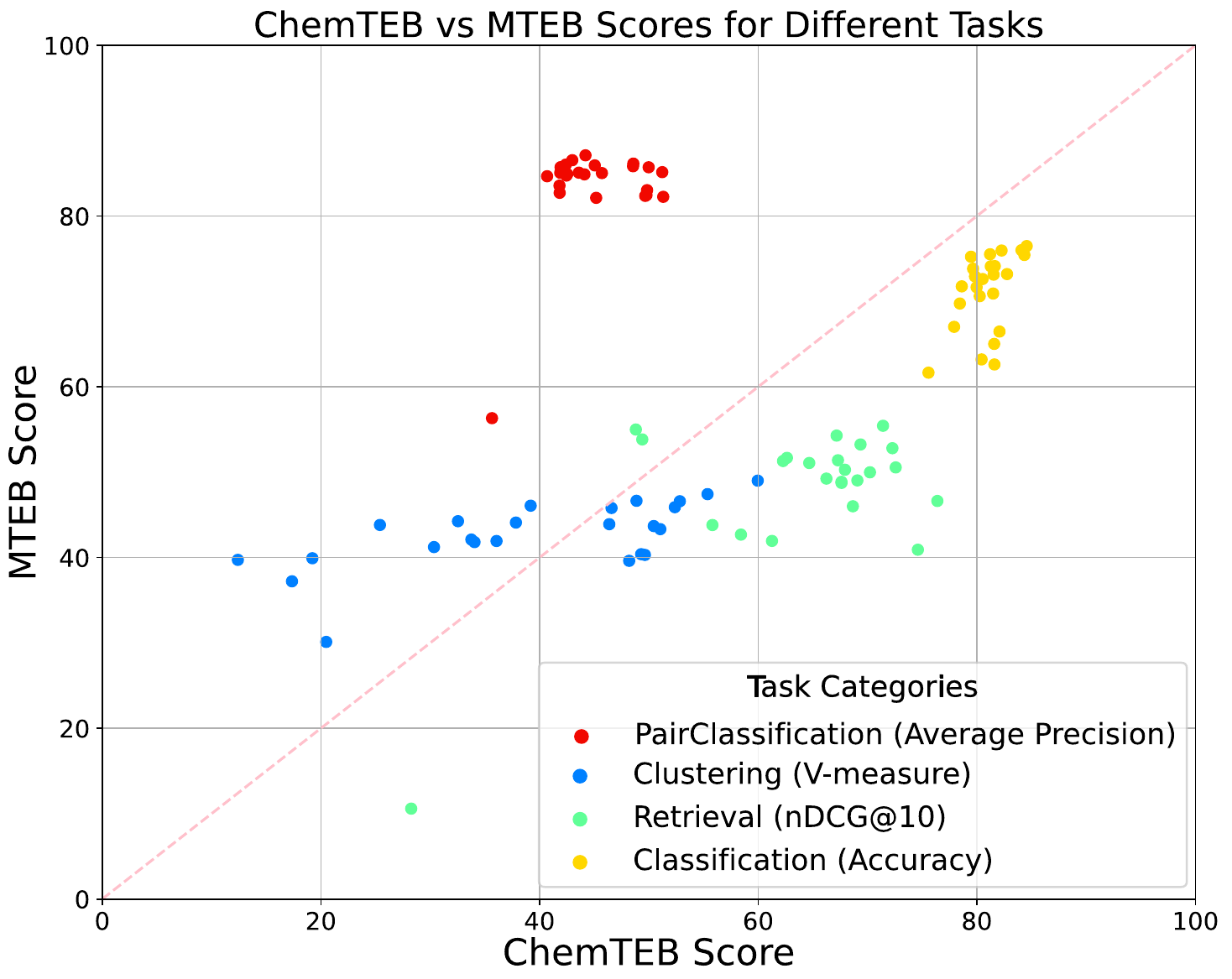}
    \caption{Comparison of model performance on ChemTEB and MTEB benchmarks across different tasks. Each point represents a model from the intersection of those tested and those on the MTEB leaderboard as of the date. The figure highlights variations in task difficulty and domain specificity.}
    \label{fig:mteb-vs-chemteb}
\end{figure}
Figure~\ref{fig:mteb-vs-chemteb} illustrates the performance of various models on both ChemTEB and MTEB benchmarks, organized by task categories. Similar metrics to the MTEB leaderboard have been used here to have comparable results. In pair classification tasks, specifically curated for chemistry, a significant decline in the average precision of models in ChemTEB tasks compared to MTEB, highlights the lack of domain expertise in evaluated models. Clustering tasks reveal that ChemTEB has more discriminating datasets, as evidenced by the variability in scores. For retrieval tasks, we observed a similar trend in terms of more variability, potentially contributed by the specialized chemical context, but the performance of each model is mostly better in ChemTEB compared to MTEB. In classification tasks, ChemTEB reflects the better performance of models with less variability in results, suggesting that these tasks may be easier compared to those in MTEB. This difference is likely due to the general nature of the Wikipedia documents used in ChemTEB classification problems. (Refer to supplementary figure~\ref{fig:mteb-vs-chemteb_supp} for details of each model performance in these two benchmarks.) 
These observations emphasize the influence of domain adaptation on model performance and underscore the necessity for tailored approaches in specialized fields like chemistry.
\section{Conclusion}
In conclusion, we have addressed a significant gap in the field of text embeddings evaluation by developing ChemTEB, a unique, open-source benchmark specifically designed for the challenges of chemical language and data. This novel tool allows for a comprehensive evaluation of both open-source and proprietary models, offering a standardized measure of performance in the chemistry domain. 
ChemTEB provides an invaluable opportunity to assess different models, identify their synergies, and pinpoint areas where innovative solutions can lead to more efficient and high-performing tools. Its model-agnostic nature ensures that it can easily evaluate any model or incorporate new data, making it a versatile addition to the open-source benchmark repertoire.

The results of our study highlight the critical importance of developing stronger, domain-adapted models for improved representation of domain-specific data in the field of chemistry.
Therefore, our work not only contributes a valuable tool to the research community but also emphasizes the need for continued innovation in the development of domain-specific models.

\begin{ack}
The author(s) gratefully acknowledge the financial support received for the research, authorship, and/or publication of this article, made possible by \textbf{MITACS} funding number IT32409. This project also benefited from the computational resources provided by the Cedar and Narval clusters of the \textbf{Digital Research Alliance of Canada}. We extend our sincere appreciation to \textbf{Adam Wojciech Bartwiki}, our project manager, whose exceptional organizational skills, effective communication, and strategic planning were invaluable in coordinating the various stages of this research project.
\end{ack}

\bibliographystyle{unsrt}
\bibliography{bib}

\newpage
\appendix

\renewcommand{\thesection}{S\arabic{section}}
\renewcommand{\thefigure}{S\arabic{figure}}
\renewcommand{\thetable}{S\arabic{table}}

\setcounter{figure}{0}
\setcounter{table}{0}
\section*{Supplementary Material}
\section{Data Sources}
\label{section:sup_data_sources}

\textbf{PubChem} \cite{10.1093/nar/gkac956} is a free, publicly accessible database of chemical molecules and their information including names, detailed descriptions, chemical formulas and properties, SMILES strings and 3D structures. We have employed PubChem to create pair classification and bitext mining datasets. One of our usages is to match SMILES strings (Isomeric or Canonical) with their corresponding entity titles and descriptions. The other usage is to match entity descriptions from different sources or their AI paraphrases to form pair classification datasets.

\textbf{Wikipedia}, with its vast array of articles, offers a comprehensive source of general and scientific knowledge, including chemistry-related content. For our specific use case, we extracted a chemistry-related subset of articles from the English Wikipedia and focused on tasks tailored to classification and clustering. Our contribution lies in creating different datasets with varying levels of difficulty and different numbers of classes, which were labeled by domain experts (chemists).

\textbf{BeIR} (Benchmarking Information Retrieval) \cite{thakur2021beir} includes datasets like HotpotQA \cite{yang2018hotpotqa} and Natural Questions (NQ) \cite{kwiatkowski2019natural} that focus on complex question-answering tasks, ideal for evaluating retrieval-based models. For our purposes, a chemistry-filtered subset of HotpotQA and NQ are leveraged for information retrieval tasks where the goal is to retrieve relevant text passages or documents in response to a query.

\textbf{CoconutDB} \cite{sorokina2021coconut} is a database of natural products, which provides comprehensive information on molecular structures and properties. This database is particularly valuable for tasks involving the analysis of natural compounds. In our approach, we focused on bitext mining and pair classification between compound formulas and their corresponding SMILES representations.

\textbf{Safety Data Sheets} utilized in this study was sourced from Kaggle \cite{msds-opp}, which compiled over 200,000 Safety Data Sheets (SDS) through web scraping. 
Following the collection phase, a portion of the data was thoroughly cleansed and annotated to enhance quality and relevance.
Our contribution lies in the creation of two specific label sets from the SDS data: \textit{$Gloves\_Required$} and \textit{$Eyes\_Protection\_Required$}. 
Notably, the majority of SDS documents indicate whether protective gloves or eye protection are necessary. To convert this unstructured text into a structured format, where the requirement for gloves or eye protection is represented as a Boolean variable (required/not required), we developed an approach that combines large language models (LLMs) with regular expression techniques. This method enabled the efficient and accurate extraction of the relevant information from the raw text.

\section{Text Embedding Models}
\label{section:supp_embed_models}
In this study, we evaluated several models on our chemical text embedding benchmark. These models are categorized into open-source and proprietary models based on their availability. Table~\ref{tab:models-summary} summarizes details about all the models that have been evaluated in this work.

\subsection{Open-source Models}

BERT \cite{devlin2018bert} introduced a groundbreaking approach to NLP by using the "masked language model" (MLM), which allows deep bidirectional context understanding by predicting masked tokens based on surrounding context. Additionally, BERT's "next sentence prediction" task enhances its ability to pre-train text-pair representations, enabling more accurate and versatile language models.

RoBERTa \cite{liu2019roberta} is a replication and improvement of BERT, addressing several limitations in BERT’s pretraining. RoBERTa introduces key modifications, including training for longer periods with larger batches, removing the next sentence prediction objective, training on longer sequences, and dynamically changing the masking pattern during training. Additionally, RoBERTa is trained on a significantly larger dataset, including the newly collected CC-NEWS corpus, which enhances its performance. These improvements allow RoBERTa to fully utilize the data, resulting in state-of-the-art performance on several GLUE tasks and matching top results on SQuAD and RACE.

SciBERT \cite{beltagy2019scibert} builds upon the architecture of BERT, leveraging the same multilayer bidirectional Transformer model but tailored for scientific text. While BERT uses WordPiece tokenization \cite{wu2016google} on general domain corpora, SciBERT constructs a specialized vocabulary (SCIVOCAB) from scientific papers using SentencePiece \cite{kudo2018sentencepiece}, resulting in a significant 42\% difference in token overlap with BERT's BASEVOCAB. SciBERT is trained on 1.14 million papers from Semantic Scholar, covering computer science and biomedical domains. The resulting model excels in scientific NLP tasks, benefiting from the specialized corpus and vocabulary.

E5 \cite{wang2022text}, short for \textbf{E}mb\textbf{E}ddings from bidir\textbf{E}ctional \textbf{E}ncoder r\textbf{E}presentations, is a general-purpose text embedding model designed to generate high-quality, single-vector representations for a wide range of tasks in both zero-shot and fine-tuned settings. Unlike models that rely on limited labeled data or low-quality synthetic text pairs, E5 is contrastively trained on CCPairs, a curated web-scale dataset that incorporates diverse data sources such as CommunityQA, Common Crawl, and Scientific papers. To maintain high data quality, CCPairs undergoes rigorous filtering using consistency-based methods. 
E5 uses a two-stage contrastive learning approach, the first stage is done with unlabeled data and in-batch negatives with a large batch size, followed by a second stage of contrastive fine-tuning on a smaller labeled dataset and hard negative mining.

Nomic AI \cite{nussbaum2024nomic} proposed open-source embedding models with a focus on increased sequence length, efficiency, and accuracy for text embedding tasks (\textit{nomic-embed-text} v1 and v1.5). It started the initial training on an improved version of BERT, called \textit{Nomic-BERT} with substituting absolute positional embeddings for rotary positional embeddings \cite{su2024roformer}, using SwiGLU activation instead of GeLU \cite{shazeer2020glu}, and using FlashAttention \cite{dao2022flashattention}. They trained the base model with a maximum sequence length of 2048 and took advantage of Dynamic NTK interpolation to scale the sequence length to 8192 for inference. Subsequently, in addition to MLM, similar to E5, Nomic employed unsupervised contrastive pre-training and supervised contrastive fine-tuning to further boost its performance in embedding tasks. 

BGE \cite{xiao2024c} leverages the RetroMAE \cite{zheng2022retromae,liu2023retromae} framework for an efficient and effective pre-training phase, which involves recovering clean text from polluted text embeddings using a lightweight decoder. Following this, BGE undergoes a two-stage training process: contrastive learning with large batch sizes and in-batch negative sampling, followed by task-specific fine-tuning. The final phase incorporates instruction-based fine-tuning, where verbal prompts guide the model to better accommodate a variety of tasks. This multi-stage training pipeline ensures that BGE not only excels in general-purpose text embedding but also in specialized applications.

The M3-embedding \cite{multim3} introduces several novel techniques to further enhance text embedding performance. One of the key innovations is self-knowledge distillation, which combines multiple outputs from different retrieval modes as a reward signal to boost the performance of single modes, particularly for sparse retrieval and multi-vector (ColBERT) retrieval. Additionally, M3 embedding improves efficiency when fine-tuning on long text through an efficient batching strategy. This small-batch approach is both simple and effective, and it can be applied to fine-tune large embedding models. Furthermore, the MCLS (Multiple CLS) method offers a straightforward way to improve performance on long texts without the need for fine-tuning, making it particularly useful for scenarios where resources for fine-tuning are limited.

\subsection{Proprietary Models}
The following proprietary models were used in our benchmarks: OpenAI's \textit{text-embedding-ada-002}, \textit{text-embedding-3-large}, and \textit{text-embedding-3-small}; Amazon's \textit{amazon.titan-embed-text-v1} and \textit{amazon.titan-embed-text-v2:0}; and Cohere's \textit{cohere.embed-english-v3} and \textit{cohere.embed-multilingual-v3}, accessed through Amazon Bedrock.

\begin{table}[ht!]
\centering
\vspace{0.75em}
\caption{This table summarizes the embedding models, highlighting each model's name, HuggingFace model or proprietary ID, model size on disk, number of parameters, the maximum context length, and the default embedding dimension. Models are categorized into open-source and proprietary sections for easier distinction.}
\vspace{0.75em}
\renewcommand{\arraystretch}{1.25}
\resizebox{\textwidth}{!}{
\begin{tabular}{lllcccc}
\toprule
    & \multicolumn{1}{c}{Model Name}     & HuggingFace Model / Model ID (Proprietary)   & Model Size           & \# Parameters        & Context length     &     Embedding size  \\
   \hline
   &  \multicolumn{5}{l}{\textbf{Open-Source Models}}                             \\
   \cmidrule{2-3}
1  & BERT                               & google-bert/bert-base-uncased                    & 440 MB               & 109.4 M              &  512             & 768     \\
2   & SciBERT                            & allenai/scibert\_scivocab\_uncased               & 442 MB               & 109.9 M              & 512          & 768          \\
3   & MatSciBERT                         & m3rg-iitd/matscibert                             & 440 MB               & 109.9 M              & 512                 & 768   \\
4   & Chemical BERT                      & recobo/chemical-bert-uncased                     & 440 MB               & 109.9 M              & 512                 & 768   \\
5   & Nomic BERT                         & nomic-ai/nomic-bert-2048                         & 549 MB               & 136.7 M              & 2048                & 768   \\
6   & Nomic Embedding v1                 & nomic-ai/nomic-embed-text-v1                     & 547 MB               & 136.7 M              & 8192                 & 768  \\
7   & Nomic Embedding v1.5               & nomic-ai/nomic-embed-text-v1.5                   & 547 MB               & 136.7 M              & 8192               & 768    \\
8   & SBERT - all Mini LM L6.v2          & sentence-transformers/all-MiniLM-L6-v2           & 90.9 MB              & 22.7 M               & 512              &  384    \\
9   & SBERT - all Mini LM L12.v2         & sentence-transformers/all-MiniLM-L12-v2          & 133 MB               & 33.3 M               & 512              & 384     \\
10  & SBERT - all MPNET-base.v2          & sentence-transformers/all-mpnet-base-v2          & 438 MB               & 109.4 M              & 514                & 768   \\
11  & SBERT - multi-qa-mpnet-base.v1     & sentence-transformers/multi-qa-mpnet-base-dot-v1 & 438 MB               & 109.4 M              & 512                & 768   \\
12  & E5 - small                         & intfloat/e5-small                                & 133 MB               & 33.3 M               & 512                 &384  \\
13 & E5 - base                          & intfloat/e5-base                                 & 438 MB               & 109.4 M              & 512                 & 768   \\
14 & E5 - large                         & intfloat/e5-large                                & 1.34 GB              & 335.1 M              & 512                & 1024  \\
15 & E5 - small v2                      & intfloat/e5-small-v2                             & 133 MB               & 33.6 M               & 512               & 384   \\
16 & E5 - base v2                       & intfloat/e5-base-v2                              & 438 MB               & 109.4 M              & 512                 & 768    \\
17 & E5 - large v2                      & intfloat/e5-large-v2                             & 1.34 GB              & 335.1 M              & 512               & 1024    \\
18 & E5 - Multilingual small            & intfloat/multilingual-e5-small                   & 471 MB               & 117.6 M              & 512                & 384     \\
19 & E5 - Multilingual base             & intfloat/multilingual-e5-base                    & 1.11 GB              & 278 M                & 514                     & 768   \\
20 & E5 - Multilingual large            & intfloat/multilingual-e5-large                   & 2.24 GB              & 559.8 M              & 514                & 1024    \\
21 & BGE - small en                     & BAAI/bge-small-en                                & 133 MB               & 33.3 M               & 512                 & 384    \\
22 & BGE - base en                      & BAAI/bge-base-en                                 & 438 MB               & 109.4 M              & 512                  & 768    \\
23 & BGE - large en                     & BAAI/bge-large-en                                & 1.34 GB              & 335.1 M              & 512              & 1024     \\
24 & BGE - small en v1.5                & BAAI/bge-small-en-v1.5                           & 133 MB               & 33.3 M               & 512                 & 384   \\
25 & BGE - base en v1.5                 & BAAI/bge-base-en-v1.5                            & 438 MB               & 109.4 M              & 512                & 768    \\
26 & BGE - large en v1.5                & BAAI/bge-large-en-v1.5                           & 1.34 GB              & 335.1 M              & 512                & 1024   \\
27 & BGE - Multilingual - M3            & BAAI/bge-m3                                      & 2.27 GB              & 576.7 M              & 8192             & 1024     \\
   &  \multicolumn{5}{l}{\textbf{Proprietary Models}}                     \\
\cmidrule{2-3}
28 & OpenAI - Text embedding 3 - small  & text-embedding-3-small                           & N/A                  & N/A                  & 8191         & 1536        \\
29 & OpenAI - Text embedding 3 - large  & text-embedding-3-large                           & N/A                  & N/A                  & 8191           & 3072      \\
30 & OpenAI - Text embedding - Ada - 02 & text-embedding-ada-002                           & N/A                  & N/A                  & 8191       & 1536           \\
31 & Amazon - Titan Text Embedding v2      & amazon.titan-embed-text-v2:0                       & N/A                  & N/A                  & 8191     & 1536             \\
32 & Amazon - Titan Embedding G1 Text      & amazon.titan-embed-text-v1                       & N/A                  & N/A                  & 8191      & 1536            \\
33 & Cohere - Embed English V3          & cohere.embed-english-v3                         & N/A                  & N/A                  & 512             & 1024         \\
34 & Cohere - Embed Multilingual V3     & cohere.embed-multilingual-v3                    & N/A                  & N/A                  & 512     & 1024      \\
\bottomrule          
\end{tabular}
}
\label{tab:models-summary}
\end{table}

\section{More Details on Chemistry Benchmarks}
MaterialBENCH \cite{yoshitake2024materialbenchevaluatingcollegelevelmaterials}  provided a dataset of problem-answer pairs for materials science and analyzing LLMs like ChatGPT and Bard on both free-response and multiple-choice questions for evaluating generative LLMs. ChemBench  \cite{mirza2024largelanguagemodelssuperhuman} introduced a framework to evaluate LLMs' chemical knowledge and reasoning abilities across diverse subfields, finding that while the best models often outperform human chemists, they struggle with some reasoning tasks and exhibit overconfidence in their predictions.
ChemLLMBench \cite{guo2023gpt} evaluated LLMs across eight chemistry-related tasks, revealing that GPT-4 performs best but highlights various limitations and the impact of in-context learning. MatDeepLearn \cite{fung2021benchmarking} benchmarks GNNs for materials chemistry applications, identifying strengths in handling compositionally diverse datasets but also noting the high data requirements and limitations. Additionally, benchmarks such as those involving simulation of hyperparameter combinations in Materials Science Optimization Benchmark \cite{BAIRD2023109487} aim to provide efficient and accurate models for optimization tasks, emphasizing the importance of surrogate models to reduce computational overhead while maintaining realistic task complexity.
Hence, a benchmark with the capability of revealing the quality of embedding models in the representation of domain-specific chemical materials was missing in the field; helping researchers to find venues for further improvement of models, and industrial users to pick the most capable models to be used in their relevant tasks. 

\section{Supplementary Methods}
\subsection{Kernel Density Estimation}
\label{sec:sup_KDE}
\textbf{Kernel Density Estimation (KDE)} is a non-parametric method used to estimate the probability density function of a random variable. Unlike parametric methods, KDE does not assume an underlying distribution for the data. Instead, it smooths the data points by placing a kernel, typically a Gaussian, at each data point. The sum of these kernels provides an estimate of the overall distribution. The KDE is controlled by a bandwidth parameter, which determines the width of the kernels and influences the smoothness of the resulting density estimate. A smaller bandwidth produces a more sensitive estimate that captures more detail in the data, while a larger bandwidth results in a smoother, more generalized estimate. KDE is widely used in data visualization, such as for plotting smoothed histograms, and is applicable in fields like finance, biology, and machine learning, where understanding data distributions is critical.

\section{Supplementary Results}
\subsection{Processing Time}
Table~\ref{tab:summary_latency} provides the details of processing time for all tasks on a V100 GPU machine. 

\begin{table}[h!]
\centering
\captionsetup{justification=centering}
\caption{Average time (seconds) to run a benchmark for each task in each category on an Nvidia V100 32GB GPU instance.}
\label{tab:summary_latency}

\resizebox{\textwidth}{!}{

\begin{tabular}{llllll}
\toprule
 & Classification & Bitext Mining & Retrieval & Clustering & Pair Classification \\
\midrule
BERT & 13.78±3.47 & 23.24±1.92 & 56.6±0.69 & 5.92±0.57 & 6.94±0.84 \\
SciBERT & 11.37±2.57 & 22.42±1.8 & 54.19±0.63 & 5.83±0.54 & 4.76±0.39 \\
MatSciBERT & 11.05±2.56 & 22.29±1.74 & 54.37±0.61 & 5.81±0.54 & 4.92±0.42 \\
Chemical BERT & 11.47±2.59 & 22.25±1.75 & 54.73±0.57 & 5.85±0.54 & 4.92±0.42 \\
Nomic BERT & 15.14±3.62 & 29.26±2.24 & 76.75±0.91 & 8.16±0.78 & 6.97±0.64 \\
Nomic Embedding v1 & 23.13±5.03 & 31.05±2.4 & 79.45±0.88 & 12.01±1.24 & 5.91±0.41 \\
Nomic Embedding v1.5 & 22.82±4.93 & 28.39±2.12 & 79.66±0.91 & 12.09±1.24 & 5.23±0.35 \\
SBERT - all Mini LM L6.v2 & 2.36±0.52 & 9.01±0.89 & 12.83±0.3 & 1.13±0.11 & 1.73±0.14 \\
SBERT - all Mini LM L12.v2 & 2.82±0.57 & 11.73±1.08 & 16.26±0.39 & 1.17±0.1 & 1.99±0.09 \\
SBERT - all MPNET-base.v2 & 11.36±2.73 & 24.49±1.87 & 61.51±0.76 & 6.26±0.6 & 4.43±0.29 \\
SBERT - multi-qa-mpnet-base.v1 & 13.29±3.09 & 24.06±1.92 & 62.06±0.67 & 7.13±0.69 & 4.42±0.3 \\
E5 - small & 4.98±1.06 & 12.55±1.11 & 21.79±0.29 & 2.36±0.22 & 2.54±0.2 \\
E5 - base & 11.24±2.67 & 23.93±1.97 & 58.84±0.75 & 6.28±0.6 & 5.28±0.46 \\
E5 - large & 37.37±9.5 & 62.4±4.78 & 191.41±2.06 & 20.46±1.99 & 14.83±1.43 \\
E5 - small v2 & 5.34±1.08 & 12.63±1.12 & 22.15±0.25 & 2.45±0.23 & 2.39±0.21 \\
E5 - base v2 & 11.21±2.66 & 24.27±2.02 & 59.45±0.73 & 6.34±0.6 & 4.83±0.49 \\
E5 - large v2 & 36.87±9.28 & 64.27±4.96 & 193.9±2.14 & 20.54±1.97 & 14.49±1.47 \\
E5 - Multilingual small & 5.2±1.11 & 11.96±1.06 & 21.68±0.28 & 2.37±0.23 & 2.29±0.2 \\
E5 - Multilingual base & 12.51±2.99 & 23.96±1.97 & 62.13±0.65 & 6.82±0.67 & 4.74±0.48 \\
E5 - Multilingual large & 40.26±10.31 & 60.97±4.51 & 209.69±0.52 & 22.01±2.18 & 13.8±1.43 \\
BGE - small en & 5.23±1.05 & 12.46±1.1 & 21.64±0.29 & 2.32±0.22 & 2.82±0.19 \\
BGE - base en & 11.14±2.64 & 23.99±1.98 & 58.64±0.72 & 6.29±0.6 & 5.32±0.48 \\
BGE - large en & 37.04±9.33 & 62.27±4.79 & 191.56±2.06 & 20.44±1.97 & 14.89±1.44 \\
BGE - small en v1.5 & 5.28±1.05 & 12.37±1.07 & 21.83±0.25 & 2.39±0.23 & 2.68±0.19 \\
BGE - base en v1.5 & 11.14±2.63 & 23.82±1.99 & 59.08±0.8 & 6.27±0.59 & 5.27±0.46 \\
BGE - large en v1.5 & 36.57±9.12 & 62.24±4.8 & 191.63±2.14 & 20.41±1.97 & 14.85±1.43 \\
BGE - Multilingual - M3 & 1139.9±251.82 & 707.86±48.87 & 3031.81±22.43 & 640.67±75.54 & 31.82±8.61 \\
OpenAI - Text embedding 3 - small & 37.17±6.89 & 372.97±36.14 & 518.72±12.57 & 27.74±2.46 & 63.49±2.91 \\
OpenAI - Text embedding 3 - large & 62.18±11.8 & 730.16±70.34 & 1006.01±27.68 & 49.39±4.65 & 123.33±5.89 \\
OpenAI - Text embedding - Ada - 02 & 35.57±6.77 & 372.55±36.18 & 518.73±12.83 & 30.77±1.88 & 64.41±2.94 \\
Amazon - Titan Text Embedding v2 & 128.01±35.05 & 1178.06±99.02 & 1595.24±34.49 & 84.65±7.8 & 244.12±3.41 \\
Amazon - Titan Embedding G1 Text & 142.23±37.78 & 1174.83±97.29 & 1627.31±40.39 & 89.03±8.38 & 243.45±3.53 \\
Cohere - Embed English V3 & 21.21±5.64 & 83.08±5.98 & 134.29±2.48 & 13.27±1.25 & 16.65±0.89 \\
Cohere - Embed Multilingual V3 & 22.32±6.07 & 80.27±5.86 & 138.74±2.51 & 14.08±1.3 & 18.07±1.29 \\
\bottomrule
\end{tabular}
}
\end{table}

\newpage
\subsection{Ranking of models}
Detailed ranking of models on each category of tasks is provided in Table~\ref{tab:summary_rank}. The ranking is calculated based on average performance over all tasks in each category defined on them.

\begin{table}[ht!]

\centering
\caption{Summary of models rank}
\resizebox{1.1\textwidth}{!}{
\begin{tabular}{lrrrrrr}
\toprule
 & Classification & Bitext Mining & Retrieval & Clustering & Pair Classification & RRF\_Score(k=10) \\
\midrule
Nomic BERT & 34 & 33 & 34 & 27 & 34 & 0.118 \\
Chemical BERT & 33 & 30 & 32 & 33 & 31 & 0.120 \\
MatSciBERT & 32 & 31 & 33 & 28 & 32 & 0.122 \\
BERT & 30 & 34 & 30 & 29 & 33 & 0.122 \\
SciBERT & 31 & 32 & 31 & 31 & 30 & 0.122 \\
BGE - small en & 12 & 27 & 26 & 25 & 22 & 0.160 \\
E5 - small v2 & 23 & 25 & 8 & 30 & 29 & 0.165 \\
E5 - small & 25 & 21 & 9 & 34 & 24 & 0.166 \\
BGE - Multilingual - M3 & 21 & 26 & 10 & 15 & 28 & 0.176 \\
E5 - base v2 & 22 & 18 & 11 & 19 & 25 & 0.178 \\
BGE - small en v1.5 & 9 & 23 & 18 & 26 & 20 & 0.180 \\
SBERT - multi-qa-mpnet-base.v1 & 28 & 29 & 24 & 17 & 5 & 0.185 \\
BGE - base en & 16 & 13 & 22 & 16 & 19 & 0.186 \\
E5 - Multilingual large & 27 & 7 & 14 & 24 & 23 & 0.187 \\
BGE - large en & 10 & 19 & 29 & 13 & 17 & 0.191 \\
E5 - base & 20 & 14 & 12 & 22 & 15 & 0.192 \\
E5 - Multilingual base & 26 & 11 & 13 & 10 & 27 & 0.196 \\
SBERT - all Mini LM L12.v2 & 18 & 22 & 23 & 21 & 4 & 0.201 \\
E5 - Multilingual small & 29 & 16 & 1 & 32 & 26 & 0.207 \\
E5 - large v2 & 24 & 12 & 3 & 23 & 21 & 0.214 \\
BGE - base en v1.5 & 15 & 17 & 7 & 11 & 18 & 0.219 \\
BGE - large en v1.5 & 6 & 15 & 15 & 18 & 12 & 0.224 \\
Amazon - Titan Text Embedding v2 & 17 & 8 & 19 & 8 & 14 & 0.224 \\
SBERT - all Mini LM L6.v2 & 8 & 20 & 20 & 20 & 3 & 0.232 \\
SBERT - all MPNET-base.v2 & 7 & 28 & 25 & 6 & 6 & 0.239 \\
OpenAI - Text embedding 3 - small & 5 & 5 & 17 & 9 & 10 & 0.273 \\
Cohere - Embed English V3 & 2 & 24 & 28 & 2 & 8 & 0.278 \\
OpenAI - Text embedding - Ada - 02 & 11 & 2 & 16 & 4 & 16 & 0.279 \\
Cohere - Embed Multilingual V3 & 4 & 9 & 27 & 3 & 9 & 0.281 \\
Nomic Embedding v1 & 19 & 10 & 4 & 12 & 2 & 0.285 \\
Amazon - Titan Embedding G1 Text & 1 & 3 & 21 & 14 & 13 & 0.285 \\
E5 - large & 14 & 4 & 6 & 5 & 11 & 0.290 \\
Nomic Embedding v1.5 & 13 & 6 & 2 & 7 & 1 & 0.339 \\
OpenAI - Text embedding 3 - large & 3 & 1 & 5 & 1 & 7 & 0.384 \\
\bottomrule
\end{tabular}
\label{tab:summary_rank}
}
\end{table}

\newpage
\subsection{Comparison of MTEB and ChemTEB}
\begin{figure}[H]
    \centering
    \includegraphics[width=0.99
    \textwidth]{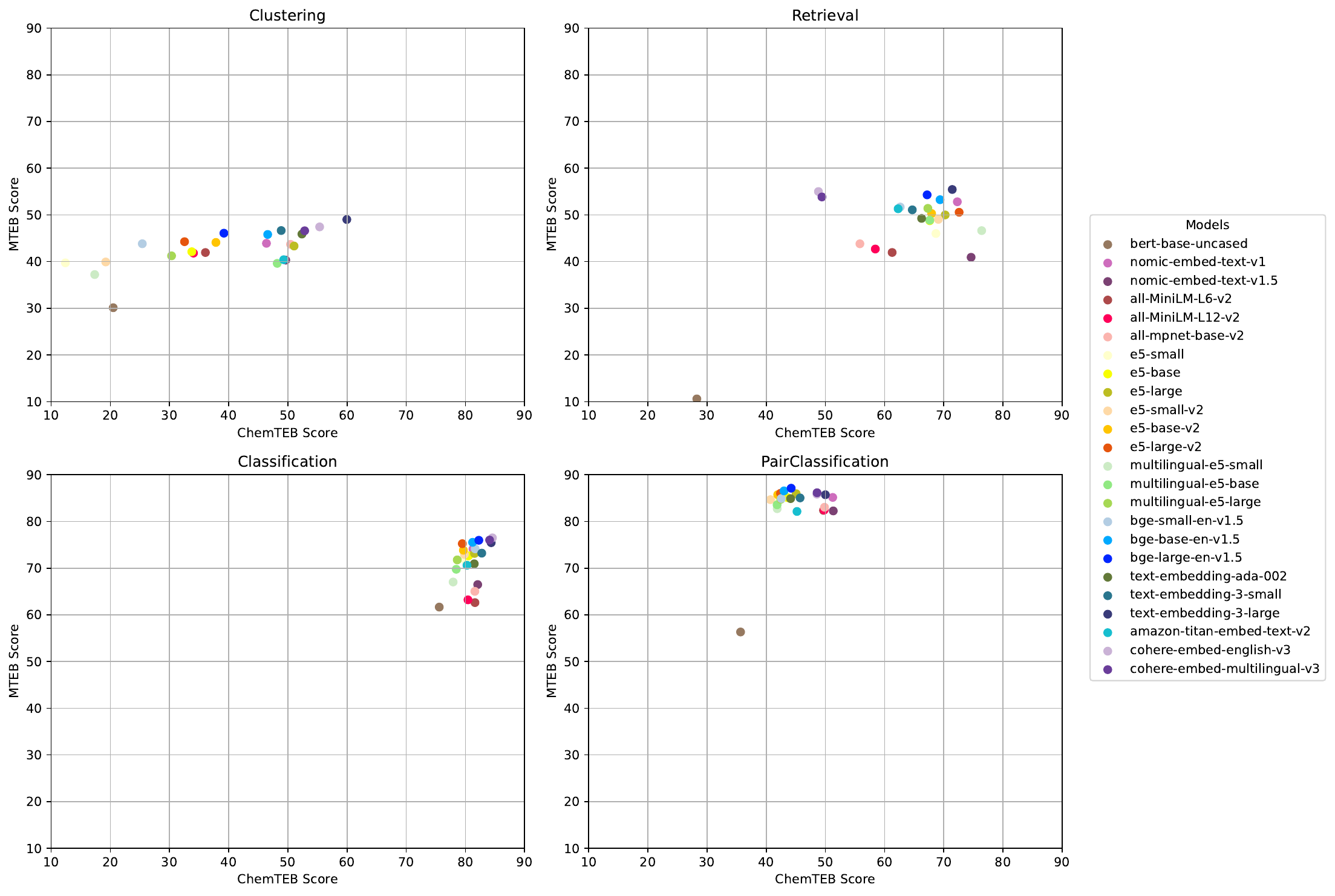}
    \caption{Comparison of model performance on ChemTEB and MTEB benchmarks across different tasks. Each point represents a model from the intersection of those tested and those on the MTEB leaderboard as of the date. The figure highlights variations in task difficulty and domain specificity.}
    \label{fig:mteb-vs-chemteb_supp}
\end{figure}

Figure~\ref{fig:mteb-vs-chemteb_supp} reflects the performance of each model in each category of tasks on both benchmarks. In three out of four categories of tasks, the BERT model provided the weakest performance in both benchmarks.

\newpage
\subsection{Correlation between models performances and tasks}
\begin{figure}[H]
    \centering
    \includegraphics[width=1.1\textwidth]{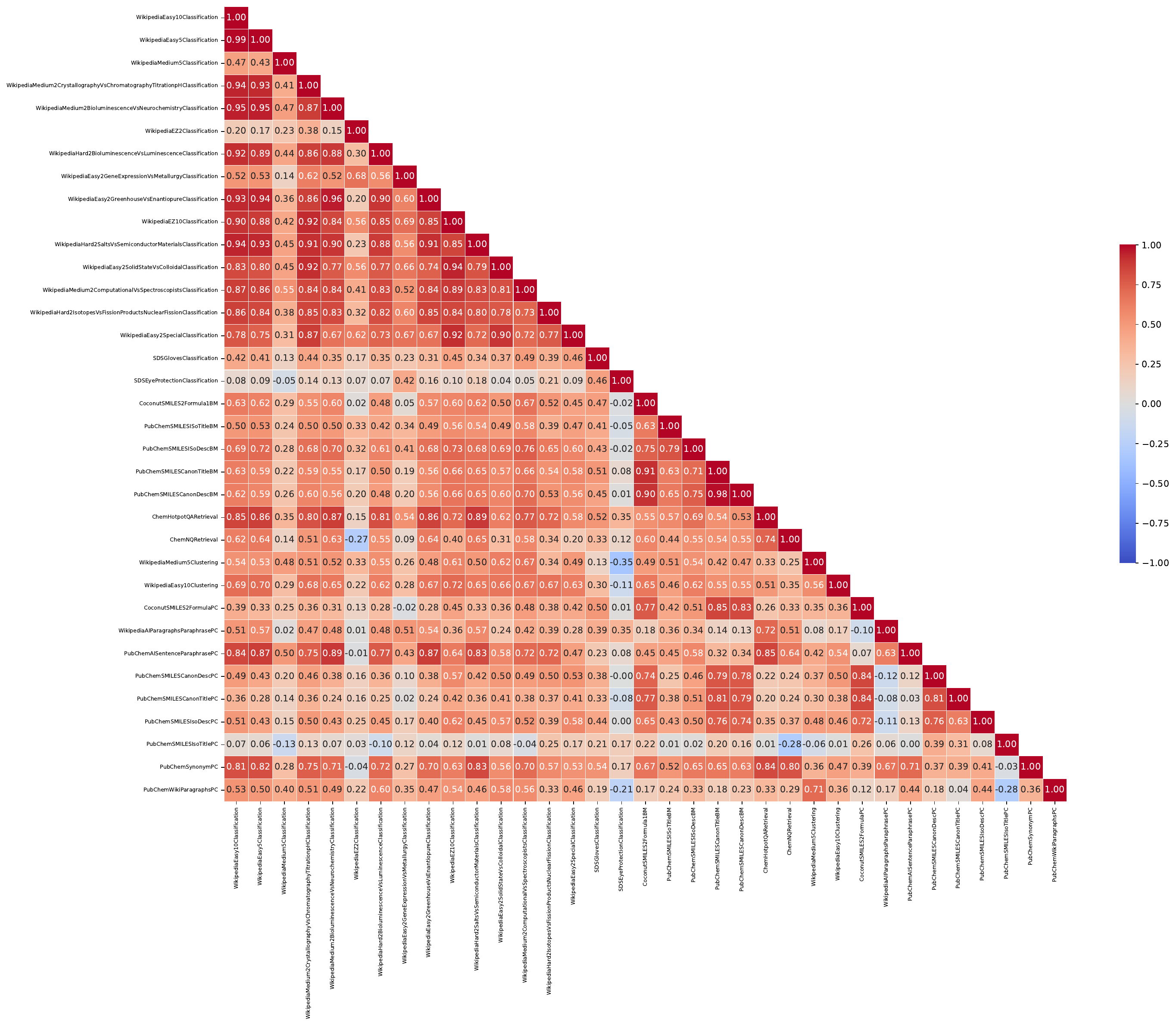}
    \caption{Correlation Matrix across datasets. Each row and column represents a separate dataset tested in the ChemTEB benchmark. The values and associated color reflect the correlation between the performance of different models on each pair of these datasets. }
    \label{fig:heatmap}
\end{figure}

\begin{figure}[H]
    \centering
    \includegraphics[width=1.1\textwidth]{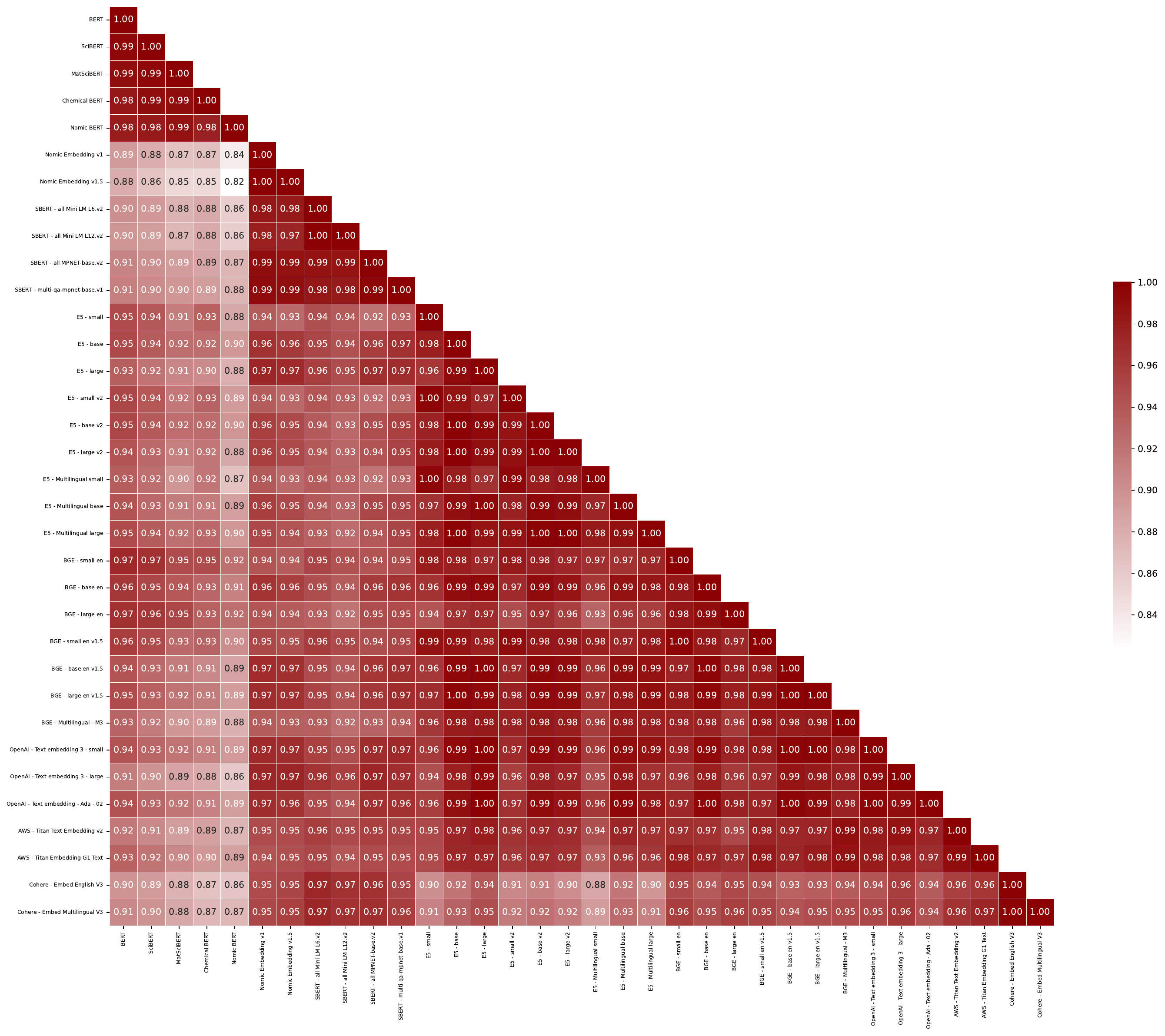}
    \caption{Correlation Matrix over Models. Each row/column represents a separate Model tested in the ChemTEB benchmark. The values and associated color reflect the correlation between the performance of each pair of models over all tested datasets.}
    \label{fig:heatmap2}
\end{figure}

Figures \ref{fig:heatmap} and \ref{fig:heatmap2} illustrate the correlation matrix for the datasets and models, respectively, with colors representing the strength of the correlations.
In figure \ref{fig:heatmap}, We can observe that in tasks such as classification, bitext mining, and retrieval, the datasets in a task are correlated except for the SDS datasets in the classification. In the clustering, and pair classification task, however, this trend is not very obvious. Especially, in the pair classification task, some of the datasets have negative correlation.

On the other hand, in figure \ref{fig:heatmap2}, we can see models categorized into two groups. Group one, the BERT-based models, and group two, the other models.
The biggest difference between these two groups is the contrastive learning being done after the pre-training, which, putting that aside the fact that BERT-based models have almost the worst performance in every single category of tasks, shows the importance of contrastive learning.
Within the second group, some model families have a stronger correlation between their performance, which may reflect closer architecture and/or pre-training tasks. For example, the Nomic embedding family has the highest correlation with the SBERT family. Cohere also reflects a closer correlation to the SBERT family first, and then the Amazon models family in the second place.   
\newpage

\end{document}